\documentclass[journal]{IEEEtran}
\usepackage{amsmath,amsfonts}
\usepackage[utf8]{inputenc}
\usepackage{algorithm}
\usepackage{array}
\usepackage[caption=false,font=normalsize,labelfont=sf,textfont=sf]{subfig}
\usepackage{textcomp}
\usepackage{stfloats}
\usepackage{url}
\usepackage{verbatim}
\usepackage{graphicx}
\usepackage{color}
\usepackage{cite}
\usepackage{algorithm}
\usepackage{algorithmicx}
\usepackage{algpseudocode}
\usepackage{amsmath}
\usepackage{multirow} 
\usepackage[top=2cm, bottom=2cm, left=2cm, right=2cm]{geometry}
\usepackage[table,xcdraw]{xcolor}
\usepackage{soul}
\usepackage{booktabs}
\usepackage{bbm}
\usepackage{amssymb}

\definecolor{MediumVioletRed}{RGB}{199, 21, 133}
\definecolor{RoyalBlue}{RGB}{65, 105, 225}
\definecolor{cBlue}{RGB}{0, 112, 192} 
\definecolor{cPink}{RGB}{192, 64, 128}
\definecolor{LYellow}{RGB}{249, 195, 50}
\definecolor{LRed}{RGB}{239, 73, 18}
\definecolor{LBlue}{RGB}{0, 176, 225}
\usepackage[colorlinks=true, citecolor=green, linkcolor=LBlue]{hyperref}

\begin{document}

\title{BAVS: Bootstrapping Audio-Visual Segmentation by Integrating Foundation Knowledge}

\author{Chen~Liu,~Peike~Li, ~Hu~Zhang,
	~Lincheng~Li, ~Zi~Huang, ~Dadong~Wang, and~Xin~Yu
\IEEEcompsocitemizethanks{
	\IEEEcompsocthanksitem C. Liu, H. Zhang, Z. Huang, and X. Yu  are with the University of Queensland, Queensland, Australia. \protect \\
	E-mail:  \{uqcliu32, hu.zhang, Helen.Huang, xin.yu\}@uq.edu.au,
    \IEEEcompsocthanksitem L. Li is with NetEase Fuxi AI Lab, Hangzhou, China. \protect\\
	E-mail: lilincheng@corp.netease.com,
	\IEEEcompsocthanksitem D. Wang is with CSIRO, Data61, Australia. \protect\\
	E-mail: Dadong.Wang@data61.csiro.au,
 	\IEEEcompsocthanksitem P. Li is with Matrix Verse. \protect\\
	E-mail: peike.li@yahoo.com.
}}

\maketitle
\begin{abstract}
Given an audio-visual pair, audio-visual segmentation (AVS) aims to locate sounding sources by predicting pixel-wise maps.
Previous methods assume that each sound component in an audio signal always has a visual counterpart in the image.
However, this assumption overlooks that off-screen sounds and background noise often contaminate the audio recordings in real-world scenarios. 
They impose significant challenges on building a consistent semantic mapping between audio and visual signals for AVS models and thus impede precise sound localization.
In this work, we propose a two-stage bootstrapping audio-visual segmentation framework by incorporating multi-modal foundation knowledge\footnote{Foundation knowledge refers to the information extracted from off-the-shelf large models.}.
In a nutshell, our BAVS is designed to eliminate the interference of background noise or off-screen sounds in segmentation by establishing the audio-visual correspondences in an explicit manner. 
In the first stage, we employ a segmentation model to localize potential sounding objects from visual data without being affected by contaminated audio signals. 
Meanwhile, %In the meanwhile, 
we also utilize a foundation audio classification model to discern audio semantics. 
Considering the audio tags provided by the audio foundation model are noisy, associating object masks with audio tags is not trivial.
Thus, in the second stage, we develop an audio-visual semantic integration strategy (AVIS) to localize the authentic-sounding objects. 
Here, we construct an audio-visual tree based on the hierarchical correspondence between sounds and object categories. %Then, we 
We then examine the label concurrency between the localized objects and classified audio tags by tracing the audio-visual tree. 
With AVIS, we can effectively segment real-sounding objects. 
Extensive experiments demonstrate the superiority of our method on AVS datasets, particularly in scenarios involving background noise.
Our project website is \url{https://yenanliu.github.io/AVSS.github.io/}.
\end{abstract}


\section{INTRODUCTION}
\label{itro}
\begin{figure}[t]
\centering
\scriptsize
\includegraphics[width=1.0\linewidth]{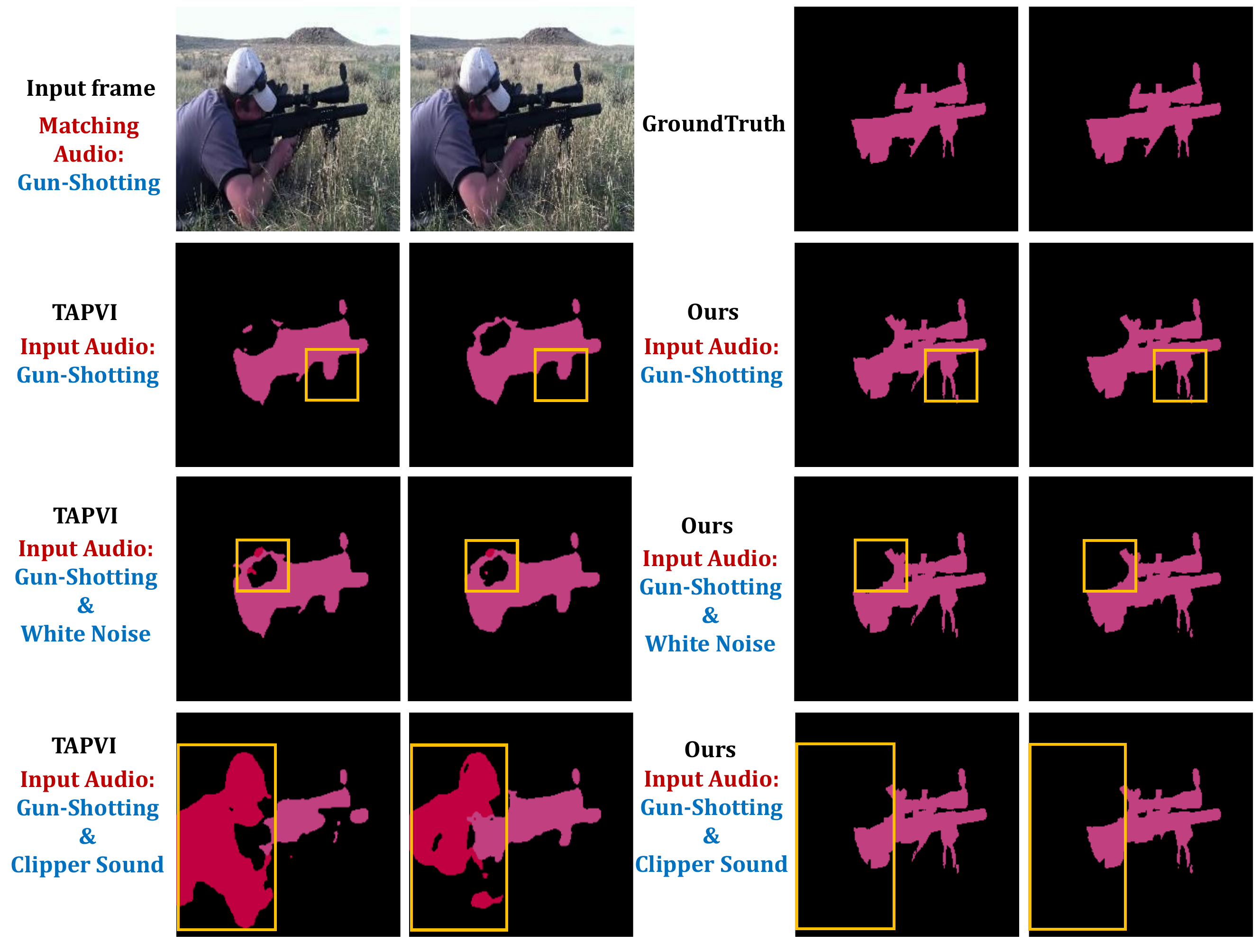}
\caption{Comparisons of our method and TPAVI \cite{zhou2023audio}. When audio signals involve background noise (\textcolor{cBlue}{white noise}) or off-screen sounds (\textcolor{cBlue}{electrical clipper}), TPAVI \cite{zhou2023audio} tends to incorporate silent regions (\emph{i.e.} \textcolor{red}{the male}) in segmentation results. In contrast, our method segments the sounding object (\textcolor{cPink}{the gun}) more precisely, especially near the gun boundaries. This implies that our method is robust against background noise and off-screen sounds. The comparison regions are highlighted by yellow bounding boxes.}
\label{fig_problemshowing}
\end{figure}

\IEEEPARstart{H}{uman} perception is multi-dimensional including vision, hearing, touching, tasting, and smell \cite{forster2011local}.
Among them, audio and visual are very important perceptual modalities in our daily life and their correspondences provide rich semantics for humans \cite{ chen2011crossmodal}. 
Understanding the correspondences fosters the development of audio-visual collaboration tasks, such as visual sound localization (VSL)~\cite{liu2022visual, mo2022closer, zhou2023exploiting} and audio-visual segmentation (AVS) \cite{zhou2022audio, zhou2023audio}. 
Unlike VSL which represents sounding objects by heatmaps, AVS strives to delineate the shape of sounding sources at a pixel level. 
 
Existing AVS methods directly decode the aligned audio-visual features into the masks of sounding sources. However, they may struggle to learn the consistent mapping between the audio-visual modalities when audio signals contain noise. 
As shown in the last two rows of Fig. \ref{fig_problemshowing}, when the input audio signal (\emph{i.e.} gun-shot) is mixed with background noise or off-screen sounds, the audio-visual segmentation results tend to incorporate silent regions.
This indicates that current AVS methods might not perform well when audios contain interference.

To tackle the above challenges, we develop a two-stage bootstrapping audio-visual segmentation framework (BAVS). 
In a nutshell, in the first stage, we intend to segment all potential sounding objects and extract the semantics of audio signals.
In the second stage, we aim to associate all potential sounding objects with the audio tags and thus attain authentic audible sources.

In the first stage, we propose to leverage a segmentation model to localize potential-sounding objects.
However, we observed that directly learning a segmentation model with the provided labels struggles to segment all potential-sounding objects.
The primary reason is that the label of a sounding object in one frame can be changed to ``background'' when it is a silent one in other visual frames.
This leads to ambiguity in learning a segmentation model and degrades the performance of the segmentation model.
To tackle this problem, we devise a silent object-aware objective (SOAO) based on the semantics extracted by an off-the-shelf large foundation model.
% senior member
To be specific, SOAO excludes penalties in segmenting silent objects that do not have ground-truth object masks in visual frames, while increasing the penalty in recognizing objects as background.
By doing so, we are able to segment various potential sounding instances.
Meanwhile, we employ a state-of-the-art audio foundation model to acquire semantic tags for each audio.
This allows us to discern the underlying audio semantics, including on-screen sounds, background noise, and off-screen sounds.

In the second stage, we focus on establishing a consistent mapping between audio and visual semantics.
One direct solution is to align the top-$k$ most confident audio tags with the corresponding segmented instance labels.
However, we found that such audio tags generated by the audio foundation model are not always accurate due to the background noise and off-screen sounds.
For example, lawn mower sounds with surrounding noise can be easily misidentified as helicopter sounds.

To overcome the aforementioned issue and localize genuine-sounding objects, we devise an audio-visual semantic integration strategy (AVIS).
Specifically, we first construct an audio-visual tree based on the hierarchical correlations between sounds and object categories, as shown in Fig.~\ref{fig_AVS_tree}.
The hierarchical structure indicates how the audio categories of potential sounds correspond to the visual categories.
% associated with each.
Moreover, to enhance the robustness of AVIS against noise tags, we group sounding sources based on the harmonics and overtones.
For instance, sounding sources such as \texttt{"car"}, \texttt{"truck"}, and \texttt{"bus"} are grouped under a parent node \texttt{"road vehicle"} as they may emit similar sounds.
In this manner, when a visual instance does not have a corresponding audio tag, we will examine whether an audio tag from the same parent node can be assigned to this instance. If we can find one, we consider this visual instance to be a genuine-sounding object.

Extensive experiments conducted in AVS datasets demonstrate that our method achieves state-of-the-art performance.
Moreover, we also illustrate that our framework is still effective and reliable in noisy audio scenarios.
Our contributions are three-fold:
\begin{itemize}
    \item We introduce a novel two-stage bootstrapping audio-visual segmentation framework (BAVS). Our framework shows strong robustness against background noise and off-screen sounds.
    \item We devise a silent object-aware objective (SOAO) to address the label-shifting issue in learning our segmentation model.
    This objective enables our model to segment all potential sounding objects.
    \item We develop an audio-visual semantic integration strategy (AVIS) by proposing an audio-visual tree. 
    This strategy allows us to establish a consistent audio-visual mapping and thus find the authentic sounding objects.
\end{itemize}

\section{RELATED WORKS}
\begin{figure*}[!t]
\centering
\includegraphics[width=0.98\linewidth]{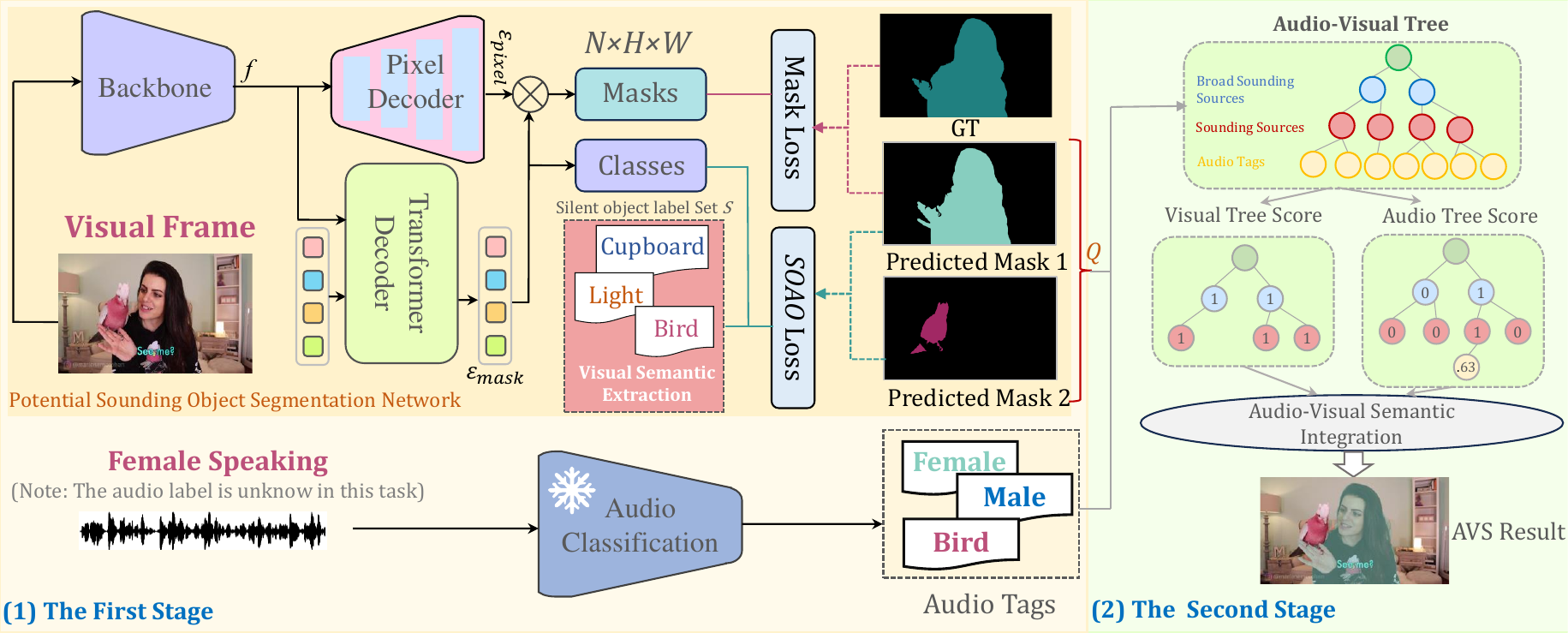} 
\vspace{-1.0em}
\caption{Overview of our BAVS framework.
In the first stage, we first utilize an off-the-shelf large foundation multi-modal model to extract the visual semantics.
Based on the visual semantics, we introduce a silent object-aware objective (SOAO) to our segmentation model and thus obtain the potential sounding instance labels and masks.
Moreover, we employ a large pre-trained audio classification foundation model to collect semantic tags for each audio recording.
In the second stage, we first introduce an audio-visual tree to fit sound semantics to their sounding sources.
Then we present the audio-visual semantic integration strategy (AVIS) to establish a consistent audio-visual mapping between the segmented instances and the audio-semantic tags.
}
\vspace{-1.5em}
\label{fig_overview_framework}
\end{figure*}

\subsection{Visual Sound Localization}
Visual sound localization (VSL) aims to localize the sounding object in the visual data based on provided audio signals \cite{senocak2018learning, oya2020we, tian2021cyclic, liu2022visual, mo2022closer, mo2022localizing, zhou2023exploiting, senocak2022less, shi2022unsupervised, song2022self, chen2021localizing}.
Predominantly, VSL methods explore the correspondence and feature similarity between audio and visual modalities to generate localization heatmaps.
For example, Senocak \emph{et al.} \cite{senocak2018learning} leverage a triplet loss to correlate audio signals with their corresponding visual frames.
Oya \emph{et al.} \cite{oya2020we} devise a similarity loss that encourages the similarity of paired audio-visual features while suppressing the similarity of non-matched pairs. 
However, these methods align the two modality features at the frame level, inevitably incorporating the silent regions during aligning.
This hinders the formation of consistent audio-visual correlations, thereby degrading the performance of VSL methods.

To enhance the localization accuracy, several works attempt to associate the sounding regions with audio signals by employing contrastive loss mechanisms \cite{chen2021localizing, mo2022closer, mo2022localizing, song2022self, senocak2022learning}.
Some approaches utilize pre-trained detection models to refine the possible sounding regions \cite{shi2022unsupervised}, while others focus on extracting challenging samples from visual frames \cite{chen2021localizing, mo2022closer}. For instance, Chen \emph{et al.} \cite{chen2021localizing} introduce a mechanism for mining challenging samples within a contrastive learning framework. 
Meanwhile, SLAVC \cite{mo2022closer} proposes a protocol for generating hard negative samples. 
However, these methods are less effective in scenarios with multiple-sounding objects since the additive nature of audio recordings.
Recently, Zhou \emph{et al.} \cite{zhou2022audio, zhou2023audio} develop audio-visual segmentation datasets, encompassing mask annotations for audible objects.
With pixel-level supervision, more precise localization can be achieved. 
In this work, we investigate the essential factors that limit the performance of audio-visual segmentation tasks and subsequently propose a two-stage framework to solve these limitations.

\subsection{Image Segmentation Networks}
\vspace{-0.2em}
Image segmentation aims to partition an image into distinct regions, where each segment represents a distinct region or a part of an object \cite{li2020consistent, li2020meta}.
Image segmentation tasks can be divided into three categories based on how semantic labels are assigned to individual pixels: semantic segmentation~\cite{ronneberger2015u}, instance segmentation~\cite{he2017mask}, and panoptic segmentation~\cite{jain2023oneformer}.
Conventionally, Deepnet-based methods are widely used to address these tasks, such as Mask-RCNN \cite{he2017mask} and U-Net \cite{ronneberger2015u}. 
Furthermore, techniques that aggregate global feature contexts, such as DANet \cite{fu2019dual}, OCNet \cite{yuan2021ocnet}, and CCNet \cite{huang2019ccnet}, have also been developed.
Recently, transformer-based architectures have gained prominence in image segmentation due to their enhanced ability to capture long-range context \cite{zheng2021rethinking, qi2023diverse, qi2023emotiongesture}.
However, these methods are tailored for only one specific sub-task, either semantic segmentation or instance segmentation. 
To address the challenge, Cheng \emph{et al.} \cite{cheng2021per} introduce a method to combine both semantic and instance segmentation tasks within a single model framework.
This pioneering methodology paves the way for sophisticated techniques that consolidate various sub-tasks within a singular architectural framework, such as Mask2Former \cite{cheng2022masked}, and MP-Former \cite{zhang2023mp}.

Different from image segmentation tasks which only consider the visual modality, audio-visual segmentation demands an intricate understanding of the interaction between audio and visual modalities.
TPAVI \cite{zhou2022audio} pioneers an approach to address the audio-visual segmentation tasks, where they first fuse features from both modalities with the cross-attention mechanism and then decode masks from the fused features.
However, this strategy impairs the effectiveness of audio semantics and amplifies visual semantic ambiguity when the audio is contaminated with noise.

\section{PROPOSED METHOD}
In this section, we present the details of the proposed framework. First, we apply the foundation models to extract prior knowledge for both visual and audio signals in Sec.~\ref{sec: Foundation Knowledge Priors}.
%two-stage bootstrapping audio-visual segmentation framework. 
We then describe potential-sounding object segmentation in Sec.~\ref{sec: Potential-sounding Objects Localization}. 
Last, we introduce audio-visual tree construction and the audio-visual semantic integration strategy (AVIS) in Sec.~\ref{sec: Audio-Visual Semantic Integration}. The overall structure of our proposed framework is illustrated in Fig.~\ref{fig_overview_framework}. 

\subsection{Foundation Knowledge Priors}
\label{sec: Foundation Knowledge Priors}
\subsubsection{Visual Foundation Knowledge Extraction}
\label{sec: Visual Foundation Knowledge Extraction}
\begin{figure}[t]
\centering
{\includegraphics[width=\linewidth]{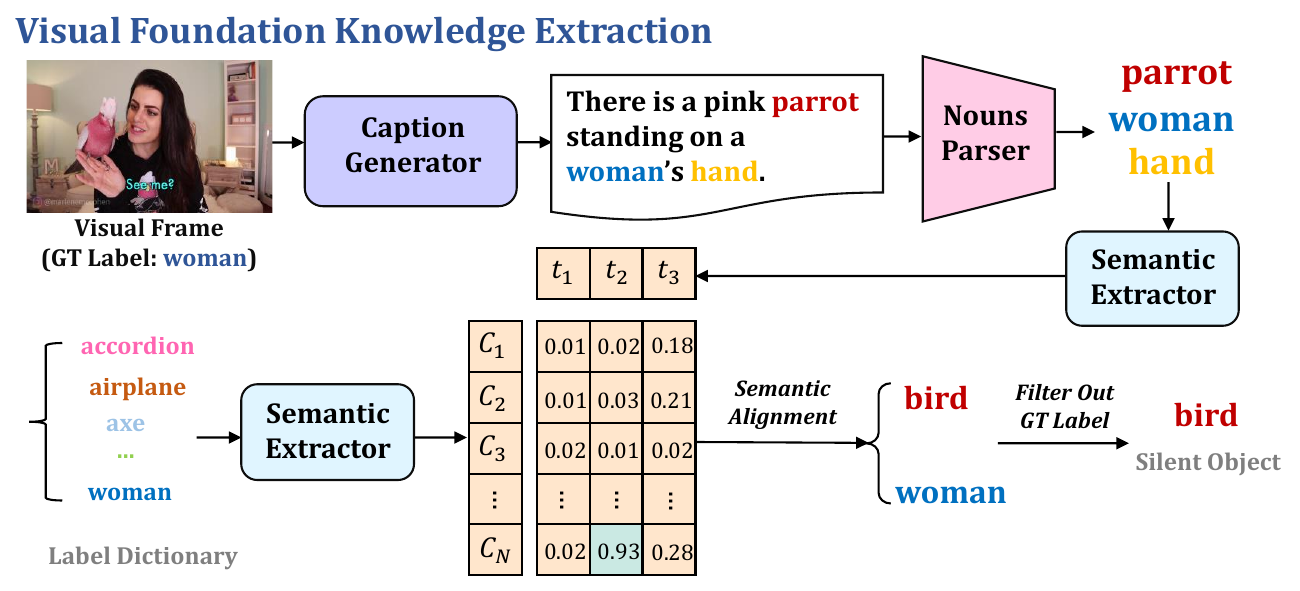}}
\caption{Illustration of visual foundation knowledge extraction. We first employ a caption generator to attain textual descriptions for the visual frame. Then we utilize the noun parser to obtain the nouns in the caption. Subsequently, a semantic extractor is adopted to obtain the representations of extracted nouns and category labels of sounding objects. We compute the similarity between extracted nouns and category labels and replace the nouns with the most similar category label. 
After filtering out the ground truth labels of the frame, we obtain the labels of silent objects.}
\label{fig_extract_semantic}
\end{figure}

To identify the semantic labels for the silent objects in each visual frame, we adopt a large off-the-shelf  foundation model (\emph{i.e.}, UniDiffuser \cite{bao2023transformer}) to extract visual foundation knowledge. Specifically, UniDiffuser is pre-trained on a large-scale dataset with paired image text.
Hence, when feeding the frames into the model, the generated captions usually contain much richer semantics and encompass an extensive vocabulary. This implies an opportunity to extract detailed semantic labels from these captions. For simplicity, we take one frame as an example to illustrate the overall process. Given the generated caption $c$ of the frame, we utilize the noun parser~\cite{bird2009natural} to collect a set of nouns $\mathcal{N} = \{n_i, i=i=1, \dots, M_1\}$, where $M_1$ denotes the number of extracted nouns. For example, as illustrated in Fig.~\ref{fig_extract_semantic}, given one generated image caption \texttt{"There is a pink parrot standing on a woman's hand"}, we can collect the nouns \texttt{"parrot"}, \texttt{"hand"}, and \texttt{"woman"} from the caption. These nouns denote the semantic labels of all potential-sounding objects in the frame, including silent objects and sounding objects.

\vspace{-0.4em}
Based on the category labels of sounding objects in the current frame, we can identify silent objects by excluding those associated with sounding objects from $\mathcal{N}$.
However, the direct exclusion is challenged by the expression inconsistency between the extracted nouns and the category labels. 
For example, UniDiffuser might extract a more specific expression such as ``parrot'', while the AVS datasets broadly categorize it as ``bird''. To address this, we implement a semantic alignment strategy, unifying different terminologies before exclusion. In conjunction with the obtained noun set $\mathcal{N}$, we characterize the category label set as $\mathcal{C} = \{c_i, i=1, \dots, M_2\}$, where $M_2$ indicates the number of object categories. Then, we employ a semantic extractor $\mathcal{F}(\cdot)$ \cite{pennington2014glove} to derive word representations for both the nouns set $\mathcal{N}$ and the category label set $\mathcal{C}$. For each noun $n_i$ in $\mathcal{N}$, we calculate its cosine similarity score $s(i, j)$ with the category name $c_j$ in $\mathcal{C}$, formulated as:
\begin{equation}
{s}(i, j) = \cos(\mathcal{F}(n_i), \mathcal{F}(c_j)),
\end{equation}
where $\cos(\cdot)$ denotes the cosine similarity function. We thus obtain a similarity score set $\mathcal{S}_i = \{s(i, k), k=1,\dots, M_2\}$ for each noun $n_i$. We sort the elements in $\mathcal{S}_i$ in descending order and identify the best matching pair $\langle n_i, c_{best}\rangle$.
The noun $n_i$ in $\mathcal{N}$ is then replaced by its corresponding category label $c_{best}$, resulting in an updated set $\mathcal{\hat{N}}= \{\hat{n}_i, i=1, \dots, M_1\}$. Leveraging set $\mathcal{\hat{N}}$, we finally obtain the semantic labels for silent objects in the current frame by excluding those encapsulated in $\mathcal{C}$. For each visual frame, we perform the same procedures and obtain their corresponding labels of silent objects.

\subsubsection{Audio Foundation Knowledge Extraction}
\label{sec: audio foundation knowledge collection}

Similarly, we employ a large off-the-shelf audio foundation model (\emph{i.e.}, Beats~\cite{chen2022beats}) to discern the audio semantics from an audio signal.
It is noteworthy that Beats~\cite{chen2022beats} is trained on AudioSet~\cite{gemmeke2017audio}, the largest audio event dataset to date. This dataset is characterized by its exceptionally detailed categorization of audio recordings, including various types of ambient noise such as wind, thunder, rain, and white noise.
Utilizing Beats enables our framework to efficiently differentiate on-screen sounds, background noise, and off-screen sounds.

In the process of obtaining audio tags for each audio signal in the AVS datasets, the first step involves forwarding the audio signal to Beats \cite{chen2022beats} to attain its audio representation. Subsequently, a projection layer is employed to generate the semantics distribution across all available categories. Lastly, a sigmoid function determines the confidence score associated with each audio tag.

\subsection{Potential-sounding Objects Segmentation}
\label{sec: Potential-sounding Objects Localization}
After obtaining the semantic labels of silent objects in Sec.~\ref{sec: Visual Foundation Knowledge Extraction}, we then introduce the details of potential-sounding object segmentation here. We first explain the details of the adopted network architecture and then elaborate on the training objectives. 

\subsubsection{Network Architecture}
\label{sec: Network Architecture}
We employ a potential-sounding object segmentation network (PSOS) to localize all potential-sounding objects. The architecture is inspired by the previous unified framework that shows excellent performance across different segmentation tasks~\cite{cheng2022masked, cheng2021per}. As indicated in Fig. \ref{fig_overview_framework}, we first feed a visual frame into a backbone network to extract its image feature $f$.
The feature is then input into a pixel decoder and a transformer decoder respectively.
The pixel decoder progressively upsamples the image feature to generate a per-pixel embedding $\varepsilon_{pixel}$, and the transformer decoder produces $N$ segmentation embeddings $\varepsilon_{mask}$.
These embeddings contribute to the formation of $N$ class predictions, along with $N$ corresponding mask embeddings.
Here, the mask embeddings are further processed by the \textit{classes head}, assigning a class label to each mask.
Furthermore, the combination of mask embeddings and per-pixel embedding $\varepsilon_{pixel}$ is fed into the \textit{masks head} to derive binary masks.
Thus, we obtain the predicted results $Q = \{(p_i, m_i), i=1, \dots, N\}$, where $p_i \in \mathbb{R}^{C+1}$ denotes the classification score and $m_i\in \{0, 1\}^{H\times W}$ symbolizes the binary mask. Note that, a ``no object" class is introduced to represent either the background regions or the categories that fall outside the scope of AVS datasets.

In practice, $N$ is usually much larger than the number of ground-truth instances in the visual images. This ensures that all potential-sounding objects have the corresponding predictions.
In calculating the training loss, we first employ the bipartite matching algorithm to identify the best prediction for each ground truth.
After obtaining the best predictions, we calculate the training loss between the matched predictions and the ground truth for optimization.

\subsubsection{Training Objectives}
Besides the obtained prediction set $Q$, we denote the ground-truth set $Q^{gt}=\{(c_j^{gt}, m_{j}^{gt}), j=1, \dots, N^{gt}\}$, where $N^{gt}$ is the number of ground truth, $c_j^{gt}$ and $m_{j}^{gt}$ represent the label and the binary mask for $j$-th ground truth, respectively.
% To train the segmentation model, we utilize the cost function $-p_i(c^{gt}_j) + \mathcal{L}_{mask}(m_i, m^{gt}_j)$ to construct the matching index set $\sigma$ between $Q$ and $Q^{gt}$.
Based on the matching index $\sigma$ between ground truth and predicted results, previous works~\cite{cheng2021per} adopt the focal loss $\mathcal{L}_{\rm focal}$ \cite{lin2017focal}, the dice loss $\mathcal{L}_{\rm dice}$\cite{milletari2016v}, and the cross entropy loss as the training objective. Mathematically, it is represented by:
\begin{equation}
\begin{aligned}
\mathcal{L}_{\rm seg} = &\sum\nolimits_{j=1}^{N} [\lambda_{f}\mathcal{L}_{\rm focal}(m_{\sigma(j)}, m^{gt}_j)\\ &+ \lambda_{d}\mathcal{L}_{\rm dice}(m_{\sigma(j)} , m^{gt}_j) -\log p_{\sigma(j)}(c^{gt}_j)],
\end{aligned}
\end{equation}
where $\lambda_{f}$ and $\lambda_{d}$ are the hyper-parameters to balance the focal loss, dice loss, and cross-entropy loss. 

Although the objective has shown good performance in traditional segmentation tasks, they fall short in AVS tasks. This primarily attributes to the label shift phenomenon inherent to the AVS dataset. Specifically, the label of an object would shift to ``background'' if the object is no longer sounding in the input audio-visual pair. This leads to ambiguity for segmentation models in training and thus causes the model overfit to the salient objects in visual frames.
To address such problems, we introduce an innovative objective function, dubbed the silent object-aware objective (SOAO). SOAO aims to increase the diversity of segmented objects, guided by semantic- and instance-level constraints.
 
The first semantic-level constraint explicitly makes the segmentation model aware of the silent objects. Previously, predictions from set $Q$ can be divided into two subsets: one corresponding to the ground truth $Q^{gt}$ and the other aligning with ``no object'' $\varnothing$. In optimization, constraints are imposed on the predictions in both subsets. Such a process neglects the potential silent objects in ``no object'', which tends to compromise the performance of obtained models in AVS tasks. Building upon the semantic labels of silent objects derived in Sec.~\ref{sec: Visual Foundation Knowledge Extraction}, we introduce a more sophisticated constraint. Specifically, we exclude the predictions that align with the silent objects and only impose constraints on the remaining predictions in the ``no object'' set. These predictions with the constraints are denoted as $\varnothing_{new}$. Mathematically, the semantic-level constraint is expressed as:
\begin{equation}
\mathcal{L}_{\rm cls} = \sum\nolimits_{j=1}^{N}- \log p_{j\in \varnothing_{new}}(c^{bg}),
\end{equation}
where $c^{bg}$ represents the ``no object'' class $p_{j\in \varnothing_{new}}(c^{bg})$ denotes the probability corresponding to the class.
 
The second instance-level constraint is designed to reduce the overlap between non-object regions and the foreground. Specifically, when masks are predicted to be the ``no object'' class, this constraint actively reduces their intersection with the ground truth object regions (foreground). This constraint is mathematically expressed as:
\begin{equation}
\mathcal{L}_{\rm ins} = \sum\nolimits_{j=1}^{N} \mathbbm{1}_{j \in \varnothing}\frac{m_{j} \cap (\bigcup_{k=1}^{N^{gt}}m^{gt}_k)}{m_{j} \cup (\bigcup_{k=1}^{N^{gt}}m^{gt}_k)},
\end{equation}
where $m_{j}$ denotes the $j$-th predicted mask in ``no object'' and $m^{gt}_k$ represent the $k$-th ground truth mask. $\bigcup_{k=1}^{N^{gt}}m^{gt}_k$ represents the ground truth objects regions. The term $m_{j} \cap (\bigcup_{k=1}^{N^{gt}}m^{gt}_k)$ represents the intersection between the ``no object'' region and foreground regions, meanwhile, $m_{j} \cup (\bigcup_{k=1}^{N^{gt}}m^{gt}_k)$ captures the union area of the two regions. 
By introducing this constraint, the model is directed towards focusing on silent regions, thereby enhancing the diversity of segmented results.

Overall, the objective of our segmentation is defined as follows:
\begin{equation}
    \mathcal{L}_{\rm SOAO} = \mathcal{L}_{\rm seg} + \lambda_{cls} \mathcal{L}_{\rm cls} +  \lambda_{ins} \mathcal{L}_{\rm ins},
\end{equation}
where $\lambda_{cls}$ and $\lambda_{ins}$ are the hyperparameters. With the training objective $\mathcal{L}_{\rm SOAO}$, our segmentation model overcomes the issue of overfitting to the most prominent ones and effectively identifies and localizes all potential-sounding objects in the visual frame.
 
\begin{figure}[t]
\centering
{\includegraphics[width=0.9\linewidth]{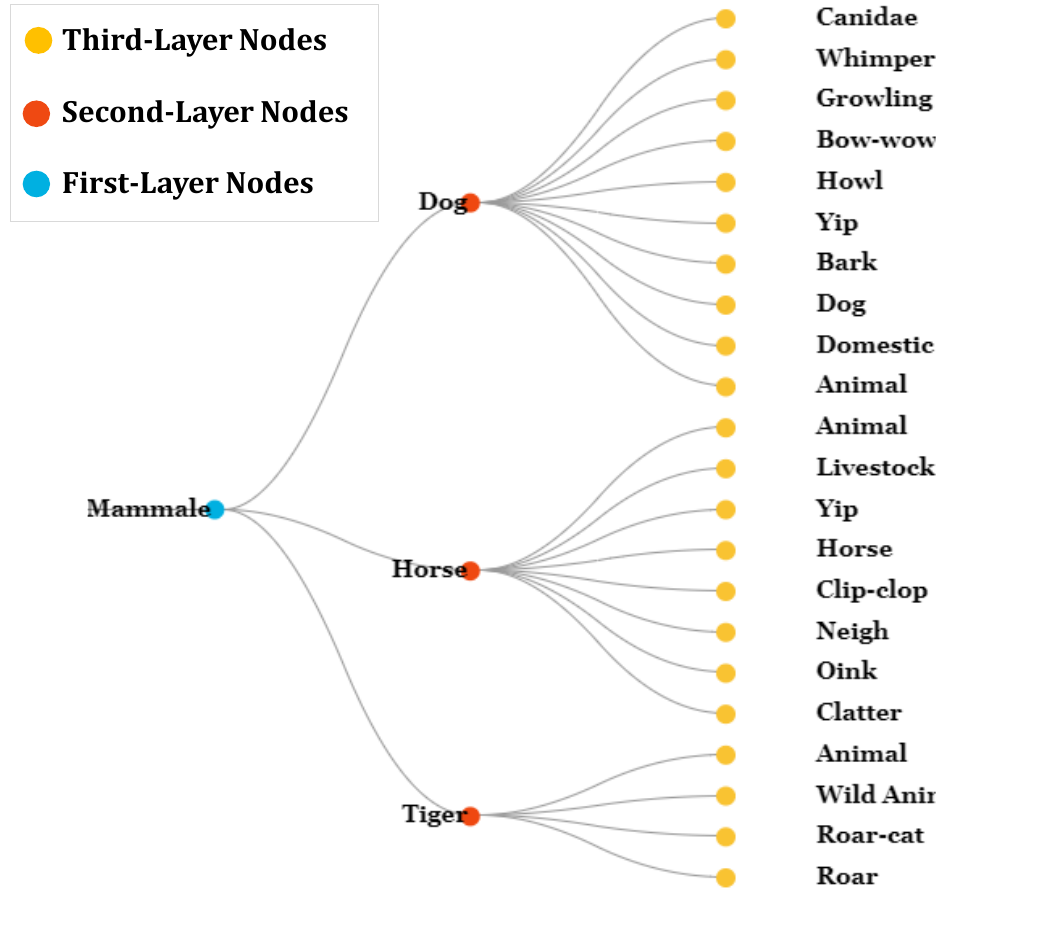}}%
\caption{
The audio-visual tree consists of three layers. In this diagram, we only display part of the nodes in each layer. The complete tree structure files can be downloaded from our project website.
From the first layer to the third layer, there are 24, 156, and 527 nodes respectively.
The \textcolor{LYellow}{yellow nodes} in the third layer represent the audio tags.
In the second layer, the \textcolor{LRed}{red nodes} are the visual categories.
Moreover, the \textcolor{LBlue}{blue nodes} are the high-level category groupings.}
\label{fig_AVS_tree}
\end{figure}

\subsection{Audio-Visual Semantic Integration}
\label{sec: Audio-Visual Semantic Integration}

\subsubsection{Audio-visual Tree Construction}
\label{sec: Audio-Visual Tree Constrcution}
To establish a consistent mapping between audio-visual semantics, we first construct an audio-visual tree.
The audio tags extracted from Beats (Sec.~\ref{sec: audio foundation knowledge collection}) are inherently granular. For example, distinct audio labels such as horse barking, clip-clop, neigh, oink, and clatter all denote horse-related sounds. Thus, we establish a mapping from these audio tags to the respective visual categories. This relationship is visualized in Fig.~\ref{fig_AVS_tree}, where each yellow node represents an audio tag, and each red node corresponds to a visual category. Using the derived confidence scores of the audio tags, we aggregate scores corresponding to the same visual category.
Any visual category (red node) with a non-zero score is then incorporated into the potential-sounding object set. This set is represented as $\mathcal{T}_c=\{t_i: \{a_j,  j=1, \dots, J\}, i=1, \dots, N_1\}$, where ${N_1}$ denotes the number of reserved nodes in this set, $\{a_j, j=1, \dots, J\}$ denotes all the audio tags that correspond to one visual category $t_i$, and $J$ is the number of audio tags. 

Considering the complexity of the input audio, the audio tags generated by the audio foundation model may be inaccurate. Specifically, the genuine categories in the audio might receive diminishing scores, while semantically analogous categories could have much higher scores. Such inaccuracy can result in the exclusion of the genuine category label from $\mathcal{T}_c$. For example, the audio of ``ambulance'' may receive a lower score for the visual category ``ambulance'' and is excluded in $\mathcal{T}_c$. Its semantically-related category ``bus'', however, might have a high score, ensuring its inclusion in $\mathcal{T}_c$. To retrieve the missed genuine category, \emph{e.g.}, ``ambulance'', 
we thus construct a high-level mapping based on the harmonics and overtones of sounds.
 %similar in semantics
If two red nodes (representing visual categories) may emit the similarity sounds, we group them in a high-level set, denoted as $\mathcal{T}_p=\{\hat{t}_i:\{t_k, k=1, \dots, K\}, i=1, \dots, N_2\}$. 
Here, $N_2$ denotes the number of nodes in the set, $\{t_k,  k=1, \dots, K\}$ denotes all the visual categories that share the similarity in one sound, and $K$ is the number of visual categories included in node $\hat{t}_i$. By integrating this mapping structure, we obtain an audio-visual tree that comprises three hierarchical layers: the third layer denotes audio tags, the second layer encapsulates visual categories of sounding objects, and the first layer signifies high-level category groupings.

\subsubsection{Audio-Visual Semantic Integration}
Based on the prediction set $Q$ from the potential-sounding object segmentation network, we also derive its potential-sounding object set for the visual data of AVS datasets. Considering the presence of numerous overlapped and low-quality masks in $Q$, we perform a two-phase filtering process. Specifically, in the first phase, we partition masks from set $Q$ into multiple subsets based on their predicted labels. 
In each subset, we arrange the objects in descending order based on their confidence score $p_i$. Masks with the highest scores are selected to form the preliminary object set $\hat{P}_c = \{(v_i, m_i), i=1, \dots, N_3\}$, where $v_i$ denotes the instance label, $m_i\in \{0, 1\}^{H\times W}$ represent the binary mask, and $N_3$ is the total number of selected objects in the first phase. In the second phase, we further identify potential-sounding objects from the remaining masks. For each remaining mask, we compute its intersection over union (IoU) scores with all the masks in $\hat{P}_c$. If all the IoU scores of the mask are below the threshold $t$, we then add this mask to $\hat{P}_c$. After performing such a process, we finally obtain the potential-sounding object set $P_c = \{(v_i, m_i), i=1, \dots, N_4\}$, where $N_4$ denotes the number of potential-sounding objects after two phases. Especially, the label $v_i$ shares the same category set with the nodes in the second layer of the audio-visual tree, ensuring consistent categorization and naming.

Based on the audio-visual tree, we perform a consistent mapping between audio-visual semantics. Specifically, for a visual mask in $P_c$, we identify it as the mask of a sounding object if its semantic label presents in both $P_c$ and $\mathcal{T}_c$. Once it is identified, the related mask and the category are removed from $P_c$ and $\mathcal{T}_c$, respectively.
If such a condition is not satisfied, we retrieve the categories that share similar sounds based on the high-level set $\mathcal{T}_p$. Similarly, if the semantic label of the mask appears both in $P_c$ and retrieved categories, we also consider it as the mask of a sounding object. In contrast, if neither condition is fulfilled, the visual mask is labeled as the mask of silent objects.

\subsection{Inference}
Given an audio-visual pair, we input the visual frame into the PSOS network to obtain the potential-sounding object set $P_c$. 
Meanwhile, we generate the audio tags for the input audio signal by leveraging the audio foundation model.
Based on the audio-visual tree, we attain the sounding source set $\mathcal{T}_c$ and high-level set $\mathcal{T}_p$ of the input audio recording.
Then, we utilize our audio-visual integration strategy to filter out the masks of silent objects in $P_c$ and obtain the final masks for sounding objects.

\section{EXPERIMENTS}
\subsection{Experimental Setup}
\begin{figure*}[t]
\centering
\scriptsize
\includegraphics[width=0.85\linewidth]{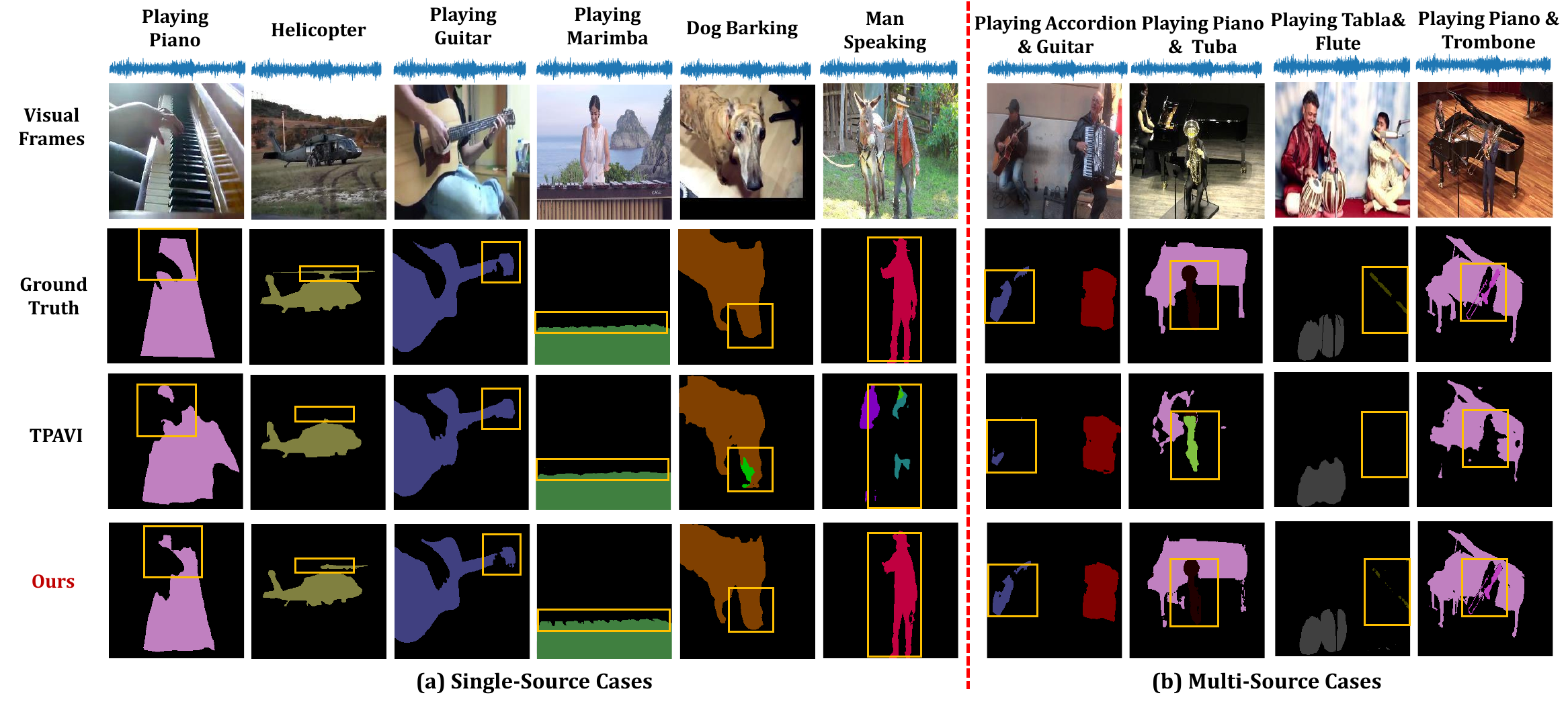}%
\vspace{-1.0em}
\caption{Qualitative comparisons with the state-of-the-art TPAVI on the single-sounding source and multi-sounding source cases. We highlight regions for comparison by yellow bounding boxes. Among all results, different colors suggest distinct sounding sources.}
\label{com_avs_fig}
\vspace{-1em}
\end{figure*}

\subsubsection{Implementation Details}
For visual masks generation, we employ Swin Transformer (Swin) \cite{liu2021swin}, Pyramid Vision Transformer v2 (PVTv2) \cite{wang2021pyramid}, and ResNet50 \cite{He_2016_CVPR} as the backbones of our segmentation network, all of which are pre-trained on ImageNet \cite{russakovsky2015imagenet}.
All visual frames are resized to a resolution of 224 $\times$ 224 $\times$ 3 pixels.
During the training stage, the models are optimized by Adam optimizer with a learning rate of 1e-4 \cite{kingma2014adam}.
We set the batch size to 20. The model with the best performance on the validation dataset is employed to segment all potential sounding objects.
For audio semantic tagging, we adopt Beats model~\cite{chen2022beats} pre-trained on Audioset \cite{gemmeke2017audio}.
The input audio signals are sampled at a rate of 48,000 samples per second, each with a duration of one second. 

\subsubsection{Audio-Visual Segmentation Datasets} 
To validate our approach, we utilize two established audio-visual segmentation datasets: AVS dataset \cite{zhou2022audio} and AVSS dataset \cite{zhou2023audio}. 
We adopt the same data split rules in~\cite{zhou2022audio} and \cite{zhou2023audio}.

\indent\textit{a) AVS Dataset:} The AVS Dataset consists of 5,356 video samples distributed over 23 categories \cite{zhou2022audio}.
In this dataset, each video is evenly split into five video clips and each clip lasts one second.
The last visual frame and the audio signal of the video clip are annotated and regarded as one audio-visual pair.
The dataset is further categorized into two distinct sub-datasets: the single-sounding source sub-dataset (AVS-S4) comprising 4,932 videos and the multi-sounding source setting (AVS-MS3) with 424 videos. 
In AVS-S4, each audio-visual pair possesses only one sounding object, while AVS-MS3 includes pairs with multiple sounding objects.
It is important to highlight that the ground truth for each pair is a binary mask, where ``0'' indicates the silent region and the ``1'' represent the sounding region.

\indent \textit{b) AVSS Dataset:} AVSS Dataset has 12,356 videos across 70 categories \cite{zhou2023audio}. 
Different from the AVS dataset which solely offers binary mask annotations, AVSS provides semantic-level annotations for sounding sources.
The dataset also introduces 7,000 new videos, each lasting 10 seconds.
Similar to the AVS dataset, each video in AVSS is uniformly divided into 10 clips, where the last frame and the corresponding audio signal form an audio-visual pair.

\renewcommand\arraystretch{1.1}
\setlength{\tabcolsep}{4.5mm}{
\begin{table}[t]
\scriptsize
\centering
\caption{Quantitative comparisons with the state-of-the-art on AVS-S4 \cite{zhou2022audio}, AVS-MS3 \cite{zhou2022audio}, and AVSS \cite{zhou2023audio}. Results of Jaccard index ($\mathcal{J}$) and the F-score ($\mathcal{F}$) are reported.}
\label{com_TPAVI}
\begin{tabular}{lcccc}
\hline
\multicolumn{1}{c}{\multirow{2}{*}{Settings}} & \multicolumn{2}{c}{TPAVI \cite{zhou2022audio}}                           & \multicolumn{2}{c}{Ours}                           \\ \cmidrule(r){2-3} \cmidrule(r){4-5}
\multicolumn{1}{c}{}                          & $\mathcal{J}$ $\uparrow$ & $\mathcal{F}$ $\uparrow$ & $\mathcal{J}$ $\uparrow$ & $\mathcal{F}$$\uparrow$ \\ \cmidrule(r){1-1} \cmidrule(r){2-3} \cmidrule(r){4-5}
AVS-S4                                & 78.74                    & 87.90                    &  \textbf{82.68}           &  \textbf{89.75}          \\
AVS-MS3                                & 54.00                    & 64.50                    &  \textbf{59.63}           &  \textbf{65.89}          \\
AVSS                                     & 29.77                    & 35.20                    &  \textbf{33.59}           &  \textbf{37.52}          \\ \hline
\end{tabular}
\vspace{-2em}
\end{table}
}

\subsubsection{Evaluation Metrics}
In this study, we employ the Jaccard Index ($\mathcal{J}$) \cite{everingham2010pascal} and the F-score ($\mathcal{F}$) to evaluate the obtained AVS models.
Specifically, $\mathcal{J}$ calculates the intersection-over-union (IoU) value between the predicted masks of sounding objects and the ground truth.
$\mathcal{F}$ provides a comprehensive evaluation of the model in terms of precision and recall. 
It is defined as $\mathcal{F}= \frac{( 1 + \beta^2)\cdot \rm precision \cdot \rm recall}{\beta^2 \cdot \rm precision + \rm recall}$. In line with previous works~\cite{zhou2022audio, zhou2023audio}, we set $\beta^2$ as 0.3.
\renewcommand\arraystretch{1.1}
\setlength{\tabcolsep}{3.0mm}{
\begin{table*}[t]
\scriptsize
\centering
\caption{Quantitative comparisons with visual sound localization methods on AVS-S4 \cite{zhou2022audio} and AVS-MS3 \cite{zhou2022audio} and AVSS \cite{zhou2023audio} datasets. For a fair comparison, we convert the semantic masks to binary masks.}
\label{com_vsl_tab}
\begin{tabular}{ccccccccccccc}
\hline
\multirow{3}{*}{Settings}   & \multicolumn{2}{c}{EZLSL \cite{mo2022localizing}}                           & \multicolumn{2}{c}{LVS \cite{chen2021localizing}}                             & \multicolumn{2}{c}{SLAVC \cite{mo2022closer}}                           & \multicolumn{2}{c}{SSPL  \cite{song2022self} w/o PCM}                     & \multicolumn{2}{c}{SSPL \cite{song2022self} w/ PCM}                               & \multicolumn{2}{c}{Ours}        \\
                            & $\mathcal{J}$ $\uparrow$ & $\mathcal{F}$ $\uparrow$ & $\mathcal{J}$ $\uparrow$ & $\mathcal{F}$ $\uparrow$ & $\mathcal{J}$ $\uparrow$ & $\mathcal{F}$ $\uparrow$ & $\mathcal{J}$ $\uparrow$ & $\mathcal{F}$ $\uparrow$ & $\mathcal{J}$ $\uparrow$ & $\mathcal{F}$ $\uparrow$ & $\mathcal{J}$ $\uparrow$ & $\mathcal{F}$ $\uparrow$ \\
                            \cmidrule(r){1-1} \cmidrule(r){2-3} \cmidrule(r){4-5} \cmidrule(r){6-7} \cmidrule(r){8-9} \cmidrule(r){10-11} \cmidrule(r){12-13}
\multicolumn{1}{l}{AVS-S4}  & 32.99                    & 44.42                    & 23.28                    & 30.83                    & 26.53                    & 35.70                    & 18.72                    & 32.52                    & 24.04                    & 36.67                   & \textbf{82.68}           & \textbf{89.75}           \\
\multicolumn{1}{l}{AVS-MS3} & 24.01                    & 27.68                    & 22.73                    & 31.59                    & 21.40                    & 23.29                    & 22.83                    & 25.23                    & 18.67                    & 23.17        & \textbf{59.63}           & \textbf{65.89}           \\ 
\multicolumn{1}{l}{AVSS}    & 30.00                    & 39.12                    & 22.07                    & 29.94                    & 28.35                    & 36.05                    & 17.04                    & 23.09                    & 15.14                    & 23.32                   & \textbf{55.45}           & \textbf{64.01}           \\ \hline
\end{tabular}
\vspace{-2em}
\end{table*}
}
\subsection{Quantitative Evaluation}
\subsubsection{Comparison with the AVS method}
We benchmark our approach against the state-of-the-art method TPAVI~\cite{zhou2022audio, zhou2023audio}. 
As suggested in Tab.~\ref{com_TPAVI}, our method consistently outperforms TPAVI on all datasets.
Note that on the AVS-MS3 sub-dataset, our method achieves a significant 5.63\% improvement with respect to metric $\mathcal{J}$.
This implies that our method excels in handling scenarios involving multiple sound sources.
Significantly, the AVSS dataset presents more challenges, characterized by more intricate audio recording environments and the necessity for semantic-level predictions.
Hence, scores under both metrics are lower when compared to those on the AVS dataset.
Nonetheless, our method still surpasses TPAVI by a large margin, \emph{e.g.}, 3.82\% with respect to metric $\mathcal{J}$.
These results demonstrate the effectiveness and robustness of our method in more complex audio-visual segmentation cases.

\begin{figure*}[t]
\scriptsize
\centering
\includegraphics[width=0.9\linewidth]{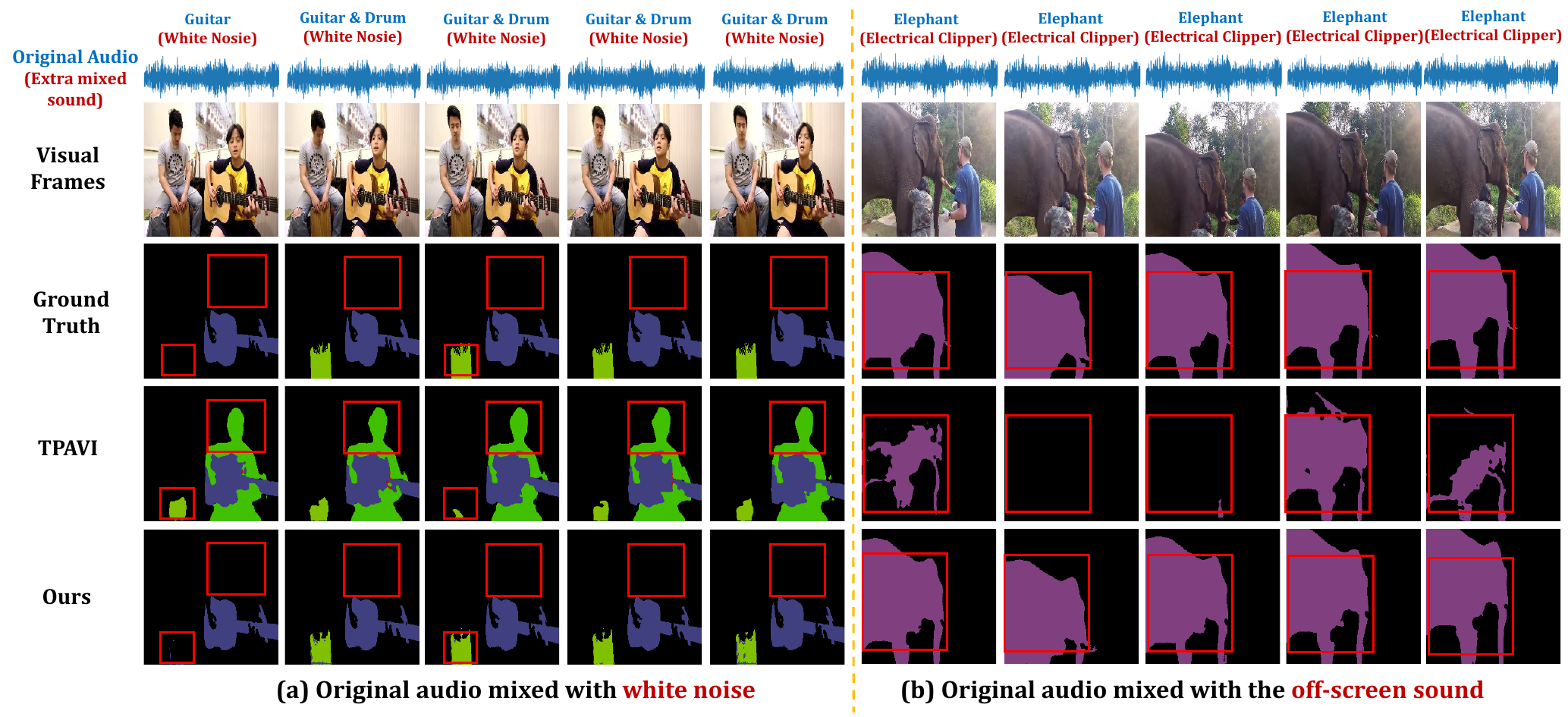}%
\vspace{-1.5em}
\caption{Visual results of adding \textcolor{red}{white noise} to \textcolor{cBlue}{original audio} recordings. Red bounding boxes highlight the specific regions for comparison. While the input audio-visual pairs are with interference, our framework can still segment sounding objects accurately.}
\label{com_noise_fig}
\vspace{-1.5em}
\end{figure*}

\subsubsection{Comparison with the VSL methods}
We further compare our method with state-of-the-art methods in Visual Sound Localization (VSL), including EZLSL \cite{mo2022localizing}, LVS \cite{chen2021localizing}, SLAVC \cite{mo2022closer}, and SSPL \cite{song2022self}. 
For a fair comparison, we re-train these models on both AVS and AVSS datasets. 
While VSL methods are crafted for locating sounding sources, their outputs are primarily heatmaps.
Hence, we cannot directly use metrics $\mathcal{F}$ and $\mathcal{J}$ to compare our method with VSL methods.
To solve this problem, we first convert the ground truth masks of the AVSS dataset into binary masks and then calculate the values under metrics $\mathcal{J}$ and $\mathcal{F}$.

As suggested by Tab. \ref{com_vsl_tab}, our method consistently outperforms the baselines under both metrics.
For example, compared to EZLSL, our approach achieves notable improvements of 49.69\% on AVS-S4, 35.62\% on AVS-MS3, and 25.45\% on AVSS with respect to the $\mathcal{J}$ metric.
It is worth noting that these VSL methods neglect the challenges introduced by off-screen sounds and background noise.
Most of them are trained on VGGSound \cite{chen2020vggsound} dataset, which comprises audio recordings curated from controlled, interference-free settings.
Although we have re-trained them on AVS and AVSS datasets, the models are still incapable of effectively handling the cases with noise or off-screen sounds due to their inherent limitations.

\subsubsection{Analysis of background noise and off-screen sounds}
To intuitively demonstrate the effectiveness of our method against background noise or off-screen sounds, we compare with TPAVI on the data with explicit background noise or off-screen sounds.
In detail, we add the white noise or the off-screen sound (\emph{i.e.}, ``electrical clipper") to each audio recording in AVSS test data. 

As suggested by Tab. \ref{com_noise}, our method exhibits superior performance across both challenging settings.
Specifically, in the case involving off-screen sounds (w/ off-screen sounds), our method witnesses a marginal drop of 0.07\% under metric $\mathcal{J}$, whereas TPAVI suffers from a significant decline of 6.21\%.
In the case of white noise (w/ noise), the $\mathcal{J}$ result of our method tends to drop from 33.59\% to 30.51\%, while TPAVI decreases from 29.77\% to 23.59\%.
These results imply that TPAVI struggles to establish a consistent mapping between audio and visual signals when facing the interference of background noise and off-screen sounds. 
In contrast, our proposed method effectively mitigates these adverse effects.

\renewcommand\arraystretch{1.1}
\setlength{\tabcolsep}{3.5 mm}{
\begin{table}[t]
\scriptsize
\centering
\caption{Impact of off-screen sounds and background noise. The experiments are conducted on AVSS \cite{zhou2023audio}. ``w/ noise'' means each audio signal in the test dataset is mixed with white noise. ``w/ off-screen sounds'' represents the original audio recordings blended with off-screen sounds.}
\label{com_noise}
\begin{tabular}{lcccc}
\hline
\multicolumn{1}{c}{\multirow{2}{*}{Setting}} & \multicolumn{2}{c}{TPAVI \cite{zhou2022audio}}     & \multicolumn{2}{c}{Ours}        \\ \cmidrule(r){2-3} \cmidrule(r){4-5}
\multicolumn{1}{c}{}                         & $\mathcal{J}$ $\uparrow$ & $\mathcal{F}$ $\uparrow$ & $\mathcal{J}$ $\uparrow$  & $\mathcal{F}$   $\uparrow$\\\cmidrule(r){1-1} \cmidrule(r){2-3} \cmidrule(r){4-5}
Original Results             & 29.77         & 35.20         & \textbf{33.59} & \textbf{37.52} \\
w/ noise                                     & 23.59         & 23.66         & \textbf{30.51} & \textbf{33.84} \\
w/ off-screen sounds                         & 23.56         & 23.62         & \textbf{33.52} & \textbf{37.43} \\ \hline
\end{tabular}
\vspace{-2em}
\end{table}
}

\subsection{Qualitative Evaluation}
\subsubsection{Comparison with the AVS method}
Fig.~\ref{com_avs_fig} illustrates the qualitative results on both single-sounding source and multi-sounding source scenarios.
In single-sounding source data~(Fig.~\ref{com_avs_fig} (a)), all methods are able to locate the sounding objects. However, our proposed method achieves more delicate results on the object boundaries.
For example, while TPAVI ignores the propeller of the helicopter, our method successfully segments that region.
\renewcommand\arraystretch{1.1}
\setlength{\tabcolsep}{0.6mm}{
\begin{table}[t]
\scriptsize
\centering
\caption{Ablation study on backbones. Results on $\mathcal{J}$ and $\mathcal{F}$ are reported. Moreover,  We provide the parameter amount (Params) and GFLOPs of AVS methods.}
\label{com_backbone_tab}
\begin{tabular}{lcccccccc}
\hline
\multirow{2}{*}{Backbone} & \multirow{2}{*}{\begin{tabular}[c]{@{}c@{}}Params\\ (MB)\end{tabular}} & \multirow{2}{*}{GFLOPs} & \multicolumn{2}{c}{AVS-S4}          & \multicolumn{2}{c}{AVS-MS3}         & \multicolumn{2}{c}{AVSS}        \\ \cmidrule(r){4-5} \cmidrule(r){6-7} \cmidrule(r){8-9}
                          &                                                                                  &                         & $\mathcal{J}$ $\uparrow$  & $\mathcal{F}$ $\uparrow$  & $\mathcal{J}$  $\uparrow$ & $\mathcal{F}$ $\uparrow$  & $\mathcal{J}$  $\uparrow$ & $\mathcal{F}$  $\uparrow$ \\ 
  \cmidrule(r){1-1}   \cmidrule(r){2-2}  \cmidrule(r){3-3}   \cmidrule(r){4-5} \cmidrule(r){6-7} \cmidrule(r){8-9}
TPAVI (ResNet)         & 91.40                                                                            & 33.88                   & 72.80          & 84.80          & 47.90          & 57.80          & 20.18          & 25.20          \\
Ours (ResNet)          & 45.52                                                                            & 6.73                    & \textbf{77.96} & \textbf{85.29} & \textbf{50.23} & \textbf{62.37} & \textbf{24.68} & \textbf{29.63} \\
  \cmidrule(r){1-1}   \cmidrule(r){2-2}  \cmidrule(r){3-3}   \cmidrule(r){4-5} \cmidrule(r){6-7} \cmidrule(r){8-9}
TPAVI (PVT)            & 101.32                                                                           & 30.28                   & 78.74          & 87.90          & 54.00          & 64.50          & 29.77          & 35.20          \\
Ours (PVT)             & 92.94                                                                            & 23.28                   & \textbf{81.96} & \textbf{88.60} & \textbf{58.63} & \textbf{65.49} & \textbf{32.64} & \textbf{36.42} \\ 
  \cmidrule(r){1-1}   \cmidrule(r){2-2}  \cmidrule(r){3-3}   \cmidrule(r){4-5} \cmidrule(r){6-7} \cmidrule(r){8-9}
Ours (Swin)               & 101.80                                                                           & 24.2                    & \textbf{82.68} & \textbf{89.75} & \textbf{59.63} & \textbf{65.89} & \textbf{33.59} & \textbf{37.52} \\ \toprule
\end{tabular}
\vspace{-2em}
\end{table}
}
In the context of multi-source cases~(Fig.~\ref{com_avs_fig} (b)), TPAVI fails to segment some sounding regions or misclassifies parts of sounding regions. These problems can be attributed to its training strategy. Specifically, TPAVI directly aligns the audio with the corresponding visual regions and neglects the multiple nuanced audio semantics in the audio recording.
This means one visual instance could be associated with multiple audio sources in one audio recording, thus resulting in impaired performance.
In contrast, our method first obtains the object masks and audio semantics, then associates them based on our proposed integration strategy. Our method avoids the problems in TPAVI and obtains high-quality segmentation masks for all sounding objects.

To show the model performance under the contaminated audio signals, we present the visualization results of our method and TPAVI in Fig.~\ref{com_noise_fig}. 
% Given different audio inputs, TPAVI generates the same segmentation masks for the same objects, indicating that TPAVI fails to 
% As suggested,  is ineffective when faced with contaminated noise.
% Its segmented results remain invariant, even given different audio inputs. In contrast, our method adapts its segmentation results in line with changes in audio signals.
For instance, when the audio signal is mixed with white noise, TPAVI tends to segment the silent region (``girl'') while our method avoids the segmentation of any silent regions (Fig.~\ref{com_noise_fig} (a)). As indicated by Fig.~\ref{com_noise_fig} (b), when the audio signal is mixed with the off-screen sound (``electrical clipper''), TPAVI tends to overlook the sounding region (``elephant") while our method still segments the intact region.
These observations highlight that our method builds strong audio-visual semantic correlations and is robust against background noise and off-screen sounds.

\begin{figure*}[t]
\centering
\includegraphics[width=0.9\linewidth]{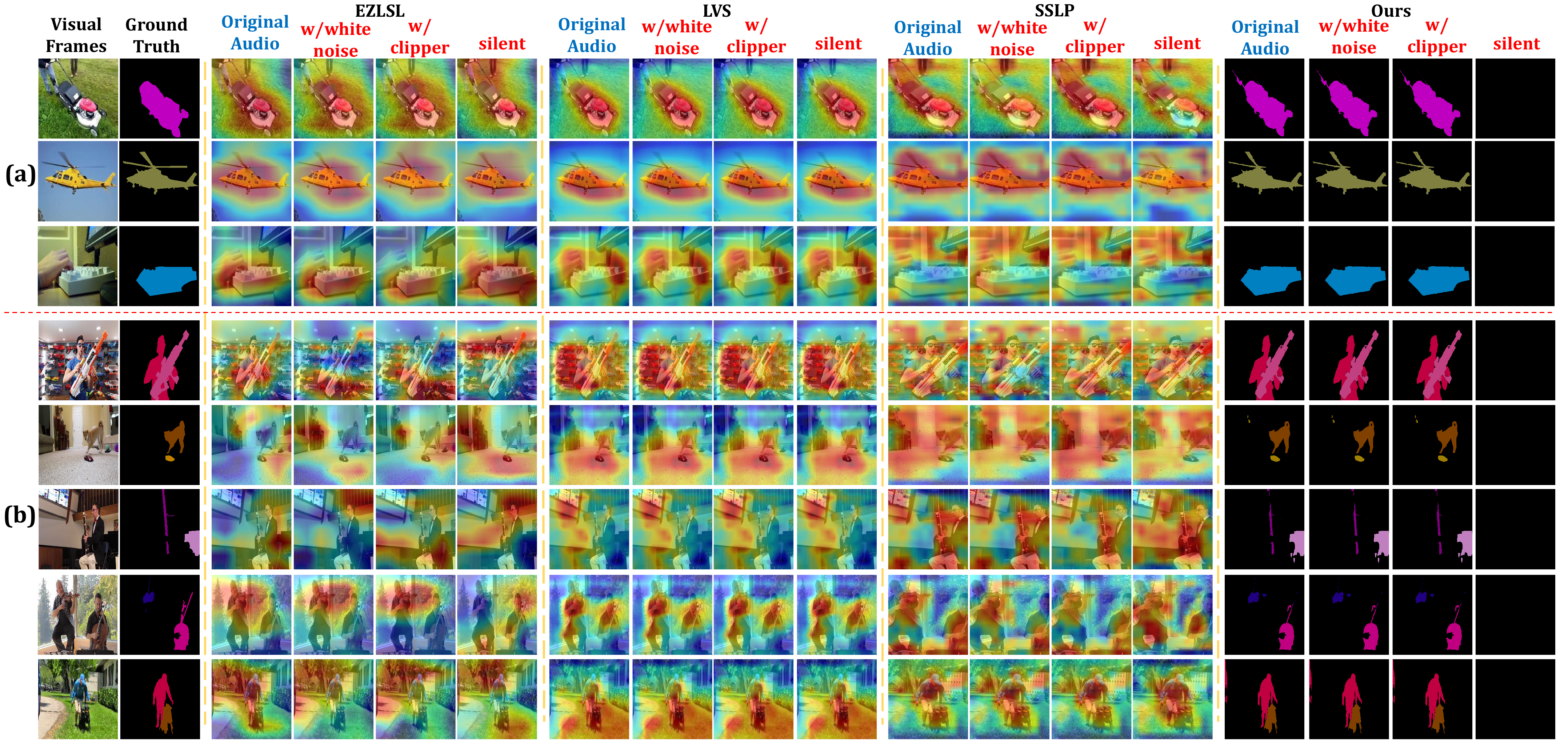}%
\vspace{-0.8em}
\caption{Visual comparisons with the visual sounding localization methods including  EZLVSL \cite{mo2022localizing}, LVS \cite{chen2021localizing}, and SSPL \cite{song2022self}. 
\textcolor{LBlue}{Original Audio} means the audio signal matches the visual frame.
(a) and (b) suggest the visual results of all methods on the single-sounding source and multi-sounding source cases, respectively.
\textcolor{red}{w/ white noise} indicates the original audio blended with the white noise, and \textcolor{red}{w/ clipper} represents the original audio mixed with the off-screen sound ``clipper''. 
\textcolor{red}{Silent} denotes that the input audio is silent.
}
\label{com_vsl_fig}
\vspace{-1.5em}
\end{figure*}

\subsubsection{Comparison with the VSL methods}
Fig. \ref{com_vsl_fig} illustrates the visual results of our method and visual sound localization methods. 
As suggested by the Fig. \ref{com_vsl_fig} (b), visual sound localization methods fail to localize the multiple sounding regions.
They tend to localize one-sounding regions or cover all significant objects within visual frames.
For instance, if the visual frame contains ``humans'', these models tend to focus only on the regions of ``humans'', even if ``humans'' do not produce any sound.
In contrast, our method avoids such problems and demonstrates a superior ability to precisely locate multiple-sounding sources.

To further illustrate the anti-interference ability of all methods, we visualize their localization results under scenarios where external sound interference or the absence of sound is considered.
As depicted in Fig. \ref{com_vsl_fig}, although the audio is silent or mixed with ``white noise'' and ``off-screen sounds'', VSL methods are still prone to localize the most prominent object in visual frames.
This observation suggests that VSL methods can not localize the genuine sounding sources but just focus on the salient objects.
In contrast, our approach demonstrates a nuanced sensitivity to audio changes and is robust against noise interference.

\renewcommand\arraystretch{1.1}
\setlength{\tabcolsep}{0.5mm}{
\begin{table}[t]
\scriptsize
\centering
\caption{Impact of the silent object-ware objective (SOAS). ``$\mathcal{L}_{cls}$" and ``$\mathcal{L}_{ins}$" are the two components of SOAS. $\mathcal{I}$ represents the number of segmented instances in the first stage.}
\vspace{-0.5em}
\label{com_soao_tab}
\begin{tabular}{lccccccccc}
\hline
\multirow{2}{*}{Setting} & \multicolumn{3}{c}{AVS-S4 \cite{zhou2022audio}}                          & \multicolumn{3}{c}{AVS-MS3 \cite{zhou2022audio}}                         & \multicolumn{3}{c}{AVSS \cite{zhou2023audio}}                         \\ 
   \cmidrule(r){2-4}  \cmidrule(r){5-7}   \cmidrule(r){8-10}
                         & $\mathcal{J}$ $\uparrow$  & $\mathcal{F}$ $\uparrow$  & $\mathcal{I}$ $\uparrow$ & $\mathcal{J}$ $\uparrow$  & $\mathcal{F}$  $\uparrow$ & $\mathcal{I}$ $\uparrow$ & $\mathcal{J}$ $\uparrow$  & $\mathcal{F}$  $\uparrow$ & $\mathcal{I}$  $\uparrow$ \\\cmidrule(r){1-1}  \cmidrule(r){2-4}  \cmidrule(r){5-7}   \cmidrule(r){8-10}
w/o  (SOAS)                  & 79.89          & 87.03          & 3700          & 53.28          & 62.71          & 320           & 23.62          & 26.59          & 11480          \\
w/ ($\mathcal{L}_{cls}$)             & 80.70          & 87.68          & 5630          & 55.36          & 63.40          & 620           & 27.31          & 30.7           & 19544          \\
w/ ($\mathcal{L}_{ins}$)             & 80.53          & 86.92          & 5564          & 54.78          & 63.01          & 576           & 27.56          & 30.92           & 20988          \\
w/ ($\mathcal{L}_{cls} + \mathcal{L}_{ins}$)        & \textbf{82.68} & \textbf{89.75} & \textbf{9962} & \textbf{59.63} & \textbf{65.89} & \textbf{910}  & \textbf{33.59} & \textbf{37.52} & \textbf{32245} \\ \hline
\end{tabular}
\vspace{-2em}
\end{table}
}
\vspace{-1.4em}
\subsection{Ablation Study}
\subsubsection{Impact of various backbones}
To demonstrate that the superiority of our method benefits from the method design rather than the backbone models, we adopt various networks, \emph{i.e.} ResNet50 \cite{He_2016_CVPR}, PVT-v2 \cite{wang2021pyramid}, and Swin \cite{liu2021swin} as the backbone for our segmentation model.
Additionally, we also provide the parameter amount (Params) and computation cost (GLOPs) of all methods for further analysis.
As suggested by Tab. \ref{com_backbone_tab}, our method consistently outperforms all other methods across different backbones.
In addition, our method attains lower Params and GFLOPs when compared with TPAVI under the same backbone.

\renewcommand\arraystretch{1.1}
\setlength{\tabcolsep}{2.5mm}{
\begin{table}[t]
\scriptsize
\centering
\caption{Impact of Audio-Visual Semantic Integration Strategy (AVIS). For the setting of \textit{w/o} AVIS, we select the instance with the highest confidence score as the predicted result.}
\vspace{-0.5em}
\label{com_AVIS_tab}
\begin{tabular}{lcccccc}
\hline
\multirow{2}{*}{Setting} & \multicolumn{2}{c}{AVS-S4 \cite{zhou2022audio}}          & \multicolumn{2}{c}{AVS-MS3 \cite{zhou2022audio}}         & \multicolumn{2}{c}{AVSS \cite{zhou2023audio}}        \\ \cmidrule(r){2-3}  \cmidrule(r){4-5}   \cmidrule(r){6-7}
                         & $\mathcal{J}$ $\uparrow$  & $\mathcal{F}$ $\uparrow$  & $\mathcal{J}$ $\uparrow$ & $\mathcal{F}$  $\uparrow$ & $\mathcal{J}$  $\uparrow$ & $\mathcal{F}$ $\uparrow$  \\ \cmidrule(r){1-1} \cmidrule(r){2-3}  \cmidrule(r){4-5}   \cmidrule(r){6-7}
w/o AVIS                 & 79.23          & 86.63          & 53.67          & 62.82          & 26.82          & 30.23          \\
w/ AVIS                  & \textbf{82.68} & \textbf{89.75} & \textbf{59.63} & \textbf{65.89} & \textbf{33.59} & \textbf{37.52} \\ \hline
\end{tabular}
\vspace{-2em}
\end{table}
}
\subsubsection{Impact of SOAO}
As discussed in Sec. \ref{sec: Potential-sounding Objects Localization}, we increase the variety of segmented masks by introducing SOAO.
To demonstrate the effectiveness of its two components ($\mathcal{L}_{cls}$ and $\mathcal{L}_{ins}$), we conduct an ablation study on the AVSS dataset. 
As suggested by Tab. \ref{com_soao_tab}, both components significantly improve the segmentation performance across all datasets.
Specifically, considering the involvement of $\mathcal{L}_{cls}$ and $\mathcal{L}_{ins}$, the obtained segmentation model achieves nearly 3\% improvement under metric $\mathcal{J}$.
Additionally, we also adopt $\mathcal{I}$ to denote the number of recalled potential-sounding objects.
The much higher performance under $\mathcal{I}$ validates that our SOAO encourages our segmentation model to segment richer objects in visual images rather than overfitting to the salient ones.

\subsubsection{Impact of AVIS}
In this work, we employ the audio-visual semantic integration strategy (AVIS) to establish the audio-visual semantic correlation.
To demonstrate the effectiveness of AVIS, we perform an ablation study on all datasets, and the results are shown in Tab. \ref{com_AVIS_tab}.
With the incorporation of AVIS, the $\mathcal{J}$ results are increased by 3.45\% on AVS-S4, 5.96\% on AVS-MS3, and 6.77\% on AVSS. The above results suggest AVIS adequately builds a strong correlation between audio and visual signals.

\section{CONCLUSIONS}
In this paper, we present a two-stage bootstrapping audio-visual segmentation framework for AVS tasks. 
Based on the visual knowledge from visual foundation models and the proposed silent object-aware objective (SOAO), our segmentation model identifies multiple potential sounding objects within the image. Meanwhile, leveraging the knowledge from the audio foundation models, we distinguish the semantics of each audio, which include the background noise and the off-screen sounds.
Finally, we use the proposed audio-visual semantic integration strategy to pinpoint the genuine sounding objects by examining the label concurrency between segmented objects and audio semantics.
Extensive experiments demonstrate the superiority of our method in addressing AVS tasks, especially the cases with background noise and off-screen sounds. We hope that our work can provide some insights to the works in the future.

\bibliographystyle{IEEEtran}
\bibliography{bibdict}
\end{document}